\documentclass{ieeeaccess}
\usepackage{cite}
\usepackage{amsmath,amssymb,amsfonts}
\usepackage{algorithmic}
\usepackage{graphicx}
\usepackage{textcomp}
\usepackage{booktabs}
\usepackage{multirow}
\usepackage{amsfonts}
\usepackage{tablefootnote}
\usepackage{comment}
\usepackage{caption}
\usepackage{bm}
\def\BibTeX{{\rm B\kern-.05em{\sc i\kern-.025em b}\kern-.08em
    T\kern-.1667em\lower.7ex\hbox{E}\kern-.125emX}}
\begin{document}
\history{Date of publication xxxx 00, 0000, date of current version xxxx 00, 0000.}
\doi{10.1109/ACCESS.2017.DOI}
\title{Investigating Persuasion Techniques in Arabic: An Empirical Study Leveraging Large Language Models}
\author{\uppercase{Abdurahmman Alzahrani*},
\uppercase{Eyad Babkier*, Faisal Yanbaawi*, Firas Yanbaawi*, and Hassan Alhuzali}}
\address{Department of Computer Science and Artificial Intelligence, Umm Al-Qura University, Makkah, Saudi Arabia.}
\tfootnote{\textbf{*} means those authors contributed equally to the work.}

\markboth
{Author \headeretal: Preparation of Papers for IEEE TRANSACTIONS and JOURNALS}
{Author \headeretal: Preparation of Papers for IEEE TRANSACTIONS and JOURNALS}

\corresp{Corresponding author: Hassan Alhuzali (e-mail: hrhuzali@uqu.edu.sa). }

\begin{abstract}


In the current era of digital communication and widespread use of social media, it is crucial to develop an understanding of persuasive techniques employed in written text. This knowledge is essential for effectively discerning accurate information and making informed decisions. To address this need, this paper presents a comprehensive empirical study focused on identifying persuasive techniques in Arabic social media content. To achieve this objective, we utilize Pre-trained Language Models (PLMs) and leverage the ArAlEval dataset, which encompasses two tasks: binary classification to determine the presence or absence of persuasion techniques, and multi-label classification to identify the specific types of techniques employed in the text. Our study explores three different learning approaches by harnessing the power of PLMs: feature extraction, fine-tuning, and prompt engineering techniques. Through extensive experimentation, we find that the fine-tuning approach yields the highest results on the aforementioned dataset, achieving an f1-micro score of 0.865 and an f1-weighted score of 0.861. Furthermore, our analysis sheds light on an interesting finding. While the performance of the GPT model is relatively lower compared to the other approaches, we have observed that by employing few-shot learning techniques, we can enhance its results by up to 20\%. This offers promising directions for future research and exploration in this topic\footnote{Upon Acceptance, the source code will be released on GitHub.}.




\end{abstract}

\begin{keywords}

Artificial Intelligence, Natural Language Processing, Persuasion Techniques Detection, Text Classification, Arabic Social Media Content.
\end{keywords}

\titlepgskip=-15pt

\maketitle

\section{Introduction} 







    
    
    

In our daily lives, we face a constant barrage of misleading information from various sources such as advertisements, radio, TV shows, and news. These sources employ persuasive techniques that can have both positive and negative effects. Understanding these techniques is crucial for navigating through the vast amount of information we encounter and making informed decisions. To shed light on the dissemination of news stories on Twitter, the Pew Research Center conducted a comprehensive study. Spanning 11 years from 2006 to 2017, the study analyzed approximately 126,000 threads that contained these stories, which were astonishingly shared a staggering 4.5 million times by around 3 million individuals. What's even more remarkable is that this rate of sharing continued to escalate beyond 2017. In today's interconnected world, where social media has significant influence, it becomes imperative to grasp the intricacies of persuasive strategies employed in text. This understanding goes beyond the content itself and encompasses the way in which the messages are delivered via the text. Consequently, there arises a pressing need to develop a tool that empowers individuals to better comprehend and evaluate the content they encounter online.

\subsection{Motivation}
Persuasive Techniques is the art of convincing, influencing and understanding the underlying psychological mechanisms that drive decision-making, persuasive techniques can range from using compelling narratives and emotional appeals to logical arguments. The effectiveness of these techniques lies  in their ability to resonate with the audience's values and beliefs, thereby motivating change in attitudes,  opinions, or behaviors. This intricate interplay between the persuader and the audience is what makes the  study and application of persuasive techniques a fundamental aspect in fields like marketing, politics, and  public relations.

Prior research investigates social media and the proliferation of persuasive techniques on people's thoughts and beliefs, as emphasized by \cite{nikolaidis2023experiments, alam2022overview,ojo2023legend,shukla2023raphael,hasanain2023araieval}, highlighting the criticality of investigating persuasive strategies in online news, aiming to shed light on manipulation attempts and enhance media literacy. By exploring various dimensions such as granularity and classification focus, it strives to provide valuable insights into the usability of automated detection methods, ultimately contributing to a more informed and vigilant online audience. 

\begin{figure}[h!]
\centering
\includegraphics[width=0.9\linewidth]{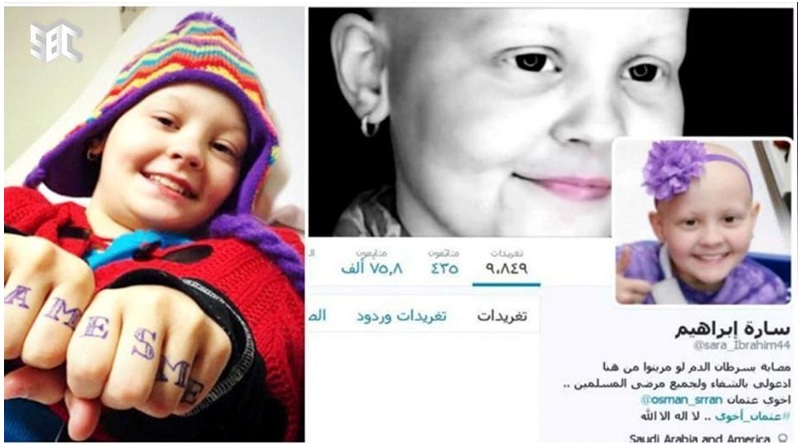}
\caption{An example of using Emotions appeal technique. Sourced from (Facebook).} \label{fig:sarah}
\end{figure}

Consider the example of Sarah Ibrahim in Saudi Arabia as shown in Figure~\ref{fig:sarah}. Sarah Ibrahim is a young girl who was widely recognized on Twitter for her battle with leukemia. People expressed their sympathy for her by sharing pictures of her undergoing treatment and requesting prayers for her recovery from this illness. Not only did social media influencers and celebrities from various fields engage with her, but they also initiated the hashtag (Friends of Sarah). To demonstrate their support, many Twitter users even posted
pictures of themselves shaving their heads. However, the shocking truth eventually emerged. It was 
revealed that Sarah Ibrahim was not a real person, but rather a fictional creation named Sarah Ibrahim. The 
child depicted in the shared images was an American child with cancer, not the Saudi girl named Sarah. 
The incident at hand serves as a clear example of employing the "appeal to emotion" technique, employed with the intention of gathering support for Sarah Ibrahim. It sheds light on the significance of verifying information and exercising prudence when interacting with online content, particularly in instances that are emotionally charged. 

\begin{figure}[h!]
    \centering
    \includegraphics[width=0.9\linewidth]{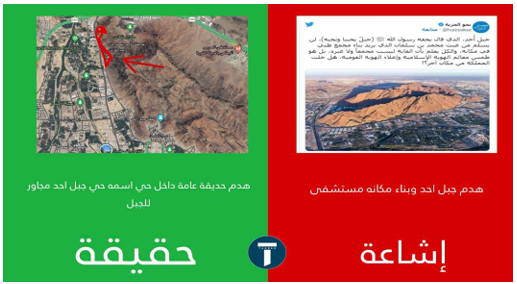}
    \caption{An example of an emotional technique used to trigger Muslims' emotion towards this event. Taken from (Tmrrah).}
    \label{fig:Mount_Uhud}
\end{figure}

Figure~\ref{fig:Mount_Uhud} shows another incident that took place on January 9, 2023, at Mount Uhud in Saudi Arabia. This incident serves as a striking illustration of how misinformation can profoundly affect society in the digital era, 
particularly when intertwined with religious sentiments. In this instance, false information quickly 
spread across online platforms, falsely asserting that the government had plans to demolish the 
sacred Mount Uhud to construct a hospital. However, the truth was that the government had only intended to demolish an adjacent park, not the actual mountain itself. The incident at Mount Uhud demonstrates the need for vigilance and critical thinking in navigating the digital landscape, where persuasion techniques can be used to effect individuals and society as a whole with inaccurate information. 

\begin{figure}[h!]
    \centering
    \includegraphics[width=1\linewidth]{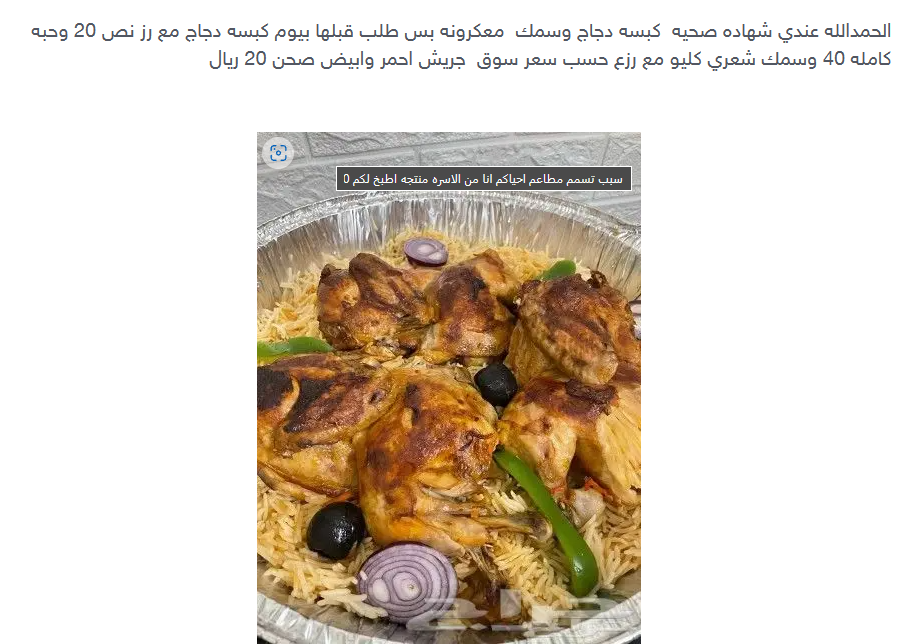}
    \caption{An example of an exploitative situation involves a food poisoning incident being used to advertise products. Taken from (Haraj).}
    \label{fig:food_poisoning}
\end{figure}

Similarly, in April 2024, a popular restaurant shut its doors following reports of food poisoning. The incident sparked a flurry of rumors on social media, quickly spreading misinformation. Unfortunately, the fallout extended beyond the affected restaurant. As shown in Figure~\ref{fig:food_poisoning}. a family posted an advertisement on Haraj website to capitalized on the situation, leveraging the incident to promote their homemade dishes. This highlights the immense impact of social media, especially in our current era. 


\subsection{Research Objective}

We discussed above the impact of using persuasive techniques in texts in terms of affecting people's emotions to proliferate false information by exploiting their feelings.
The main objective of this paper is to empirically study the use of persuasion techniques on social media by utilizing PLMs. Our investigation is specifically focused on understanding how persuasive techniques and manipulation are employed through texts. We have a particular emphasis on the dissemination of misinformation and aim to shed light on the various persuasive techniques employed in texts that are strategically designed to influence opinions and alter viewpoints. Furthermore, we address ethical considerations and potential societal impacts, including the negative use of persuasive techniques for spreading misinformation. This involves cataloging and comprehending different styles of persuasive methods used in social media, such as emotional appeals and the use of biased or misleading information. Our contributions can be summarized as follows:

\begin{itemize}
    \item Conducting a comprehensive empirical study to recognize persuasion techniques in Arabic written text.

    \item Utilizing various PLMs to categorize social media data for two tasks:
    
    \begin{itemize}
        \item Determining the presence or absence of persuasion techniques.
         \item Identifying the specific type of persuasion techniques employed.
    \end{itemize}

    \item Presenting diverse analyses to highlight the results obtained from the utilized PLMs, while also discussing the limitations of current PLMs in effectively detecting persuasion techniques.
\end{itemize}


\section{Related Work}

There's extensive research in fields like Natural Language Processing (NLP) focused on detecting persuasive techniques in text and speech~\cite{ojo2023legend,hasanain2023araieval,shukla2023raphael,nobakhtian2023iust}. This body of research aims to systematically identify and categorize persuasive tactics in different media, such as political speeches, advertisements, and social media. The primary objective is to build models that automate the detection and analysis of strategies employed in text, such as appeals to authority or straw-man effects. Such tools not only help identify how persuasion is used across different platforms but also support the development of measures against its manipulative use in spreading misinformation and propaganda. As the digital landscape evolves, maintaining ethical standards in communication becomes crucial, making this area of study increasingly vital for ensuring transparency and integrity in public discourse.

In the realm of studying persuasion techniques, one highly regarded book is authored by Cialdini \cite{cialdini1984influence}. This book focuses on describing the psychology behind people's inclination to say "yes" and offers insights on how to effectively apply this understanding. Another notable source in this field is the book by \cite{bwhayik} who attempts to provide a comprehensive exploration of various argumentative strategies, with a particular emphasis on the logical fallacy known as the 'Straw Man.' These two books serve as illustrative examples that center around the concept of persuasion techniques. 

\subsection{Pre-trained Language Models}

Significant progress has been made in developing PLMs for the English language, demonstrating their effectiveness in various Natural Language Processing (NLP) tasks. These powerful PLMs are trained on massive volumes of textual data, allowing them to perform a wide range of tasks. Prior models like ULMFiT~\cite{howard2018universal} excel in text classification tasks across languages, while BERT~\cite{devlin2019bert} established a foundation for ``deep bidirectional pre-training'', achieving SOTA performance on various sentence-level tasks. Subsequent advancements addressed limitations of BERT. ALBERT tackles memory and training time issues with parameter reduction techniques~\cite{lan2020albert} , while RoBERTa refines the pre-training process for improved performance~\cite{liu2019roberta}. XLNet goes further, addressing shortcomings in BERT's handling of masked positions and pre-training approaches~\cite{yang2019xlnet}. Moreover, Sahoo et al. \cite{sahoo2024systematic} writes a survey on prompt engineering in which they evaluate methods like zero-shot, few-shot, and chain-of-thought prompting. The authors highlight models such as PLMs and GPT-3, and they found Prompt engineering techniques, particularly with chain-of-thought prompting, excels in complex reasoning tasks. These advancements showcase the continuous development and growing capabilities of general domain PLMs for English NLP tasks.

One of the first efforts to construct a PLM specifically for Arabic was hULMonA \cite{eljundi2019hulmona}, which is the first universal language model specifically designed for Arabic. This model obtained SOTA performance in Arabic sentiment analysis tasks, demonstrating its effectiveness on various Arabic datasets. AraBERT, another early effort, pre-trained the BERT model for Arabic, achieving advanced results in sentiment analysis, entity recognition, and question answering~\cite{antoun2020arabert}. Subsequent studies explored multilingual models like GigaBERTs and focused on the impact of data sources like OSCAR and Gigaword~\cite{lan2020empirical}. ARBERT and MARBERT are two models that were built by~\cite{abdul2020arbert}, and they were trained on a large number of Modern Standard and Dialect Arabic datasets, respectively. Another model by~\cite{inoue2021interplay} pre-train a single BERT-base model called CAMeLBERT, based on variations of dialect and classic Arabic data.~\cite{almazrouei-etal-2023-alghafa} introduced ``AlGhafa'', which is focused on benchmarking Arabic PLMs for multiple-choice evaluation. An additional work by~\cite{abdelali-etal-2024-larabench} conducts benchmarking recent advancements in PLMs against state-of-the-art models on various NLP tasks.

\subsection{Resources for Persuasion Techniques Detection }

There is an existing body of literature of the topic of persuasion techniques detection. UnedMediaBiasTeam \cite{rodrigo2023unedmediabiasteam} examined genre categorization challenges and transfer learning benefits for identifying persuasive techniques in news articles. SinaAI's \cite{sadeghi2023sinaai} explored multilingual news framing, revealing language differences. ReDASPersuasion \cite{qachfar2023redaspersuasion} developed cross-lingual predictive models using language transformers to detect persuasion techniques. Volta \cite{gupta2021volta} specifically identified persuasive elements in multi-modal content, particularly memes, utilizing BERT-based models. NLyticsFKIE \cite{pritzkau2021nlyticsfkie} investigated persuasion techniques in textual and multi-modal settings, particularly in disinformation campaigns on social networks, employing RoBERTa-based architectures. Another study by Ojo et al.\cite{ojo2023legend} utilized XLM-RoBERTa as a feature extractor for detecting persuasion techniques, showcasing its versatility and effectiveness in identifying subtle persuasive strategies within Arabic tweets and news articles.

In exploring unique perspectives on persuasion, Guerini et al \cite{guerini-etal-2015-echoes} delved into the persuasive impact of phonetic features, revealing nuances in the effectiveness of euphonic sentences. Koto et al. \cite{koto-etal-2022-pretrained} investigated the role of PLMs in generating persuasive ad texts for e-commerce, considering the fluency of AI-generated texts while highlighting challenges in maintaining accuracy. Samad et al. \cite{samad2022empathetic} introduced empathetic responses within persuasive dialogue systems, enhancing their persuasiveness. Yang et al. \cite{yang2019let} presented a semi-supervised neural network model for identifying persuasive strategies, particularly in crowdfunding requests. Lastly, researchers such as chakraberty et al. \cite{chakrabarty2020ampersand} developed AMPERSAND, a model tailored for mining arguments in online discussions, enhancing the analysis of persuasive communication in digital platforms. Overall, these research studies contribute to our understanding of persuasive communication and its impact on critical thinking, media literacy, and informed decision-making in an ever-evolving digital world.

Another area of research that bears similarities to persuasion techniques is the field of fallacy detection and classification. This body of literature explores the identification and categorization of fallacies, which are flawed reasoning patterns, in various forms of communication. Fallacies can be found in written text, spoken discourse, and other mediums of communication. The study of fallacies detection and classification intersects with the study of persuasion techniques because both aim to analyze and understand the strategies used to influence or manipulate an audience. While persuasion techniques focus on effective communication strategies for influencing beliefs or behavior, fallacies detection and classification focus on identifying and analyzing flawed reasoning and logical errors. Several studies have explored the detection and classification of logical fallacies in text across various domains. Jin et al.~\cite{jin2022logical} introduced a novel task that specifically focuses on logical fallacies in text, with a subset dedicated to fallacies related to climate change. Their work aimed to identify and analyze instances of logical fallacies within textual content. A further study by Goffredo \cite{goffredo2023argument} centered on the detection of fallacies in political texts. Their approach focused on argument-based detection and classification of fallacies, utilizing transformers-based models. By leveraging these models, they aimed to identify and categorize fallacies present in political discourse. In a different vein, Alhindi \cite{alhindi2023multitask} proposed a multi-task prompting method for recognizing fallacies across various datasets and domains. This method utilized the T5 model proposed by Raffel et al.~\cite{raffel2020exploring} and aimed to detect fallacies in different contexts. These studies collectively contribute to the ongoing research on logical fallacies in text, exploring different domains, tasks, and methodologies.

\subsection{How is this work different from existing works?}

Our research differs from prior research in examining the persuasive techniques employed in Arabic language content across social media platforms. We specifically intend to explore the most effective methodology for analyzing persuasive techniques. To achieve this, we employed various NLP methods, including PLMs as feature extractor, fine-tuning PLMs, and prompt engineering PLMs. For feature extraction, we utilized PLMs such as AraBERTV2~\cite{antoun2020arabert}, MARBERT~\cite{abdul-mageed-etal-2021-arbert}, CAMelBERT~\cite{inoue-etal-2021-interplay}, and GigaBERT~\cite{lan2020gigabert}. Subsequently, we employed the same set of models for fine-tuning, along with GPT-3.5 and GPT-4~\cite{openai2024gpt4}, for prompt engineering. Through rigorous evaluation and comparison of the results, we aim to determine which approach yielded the most effective outcomes. By leveraging a diverse set of methodologies and PLMs, our research objective is to provide valuable insights into the analysis of persuasive techniques in Arabic social media content.

\section{Data and Task Settings} \label{Data and Task Settings}

\begin{figure*}[]
    \centering
\includegraphics[width=\linewidth]{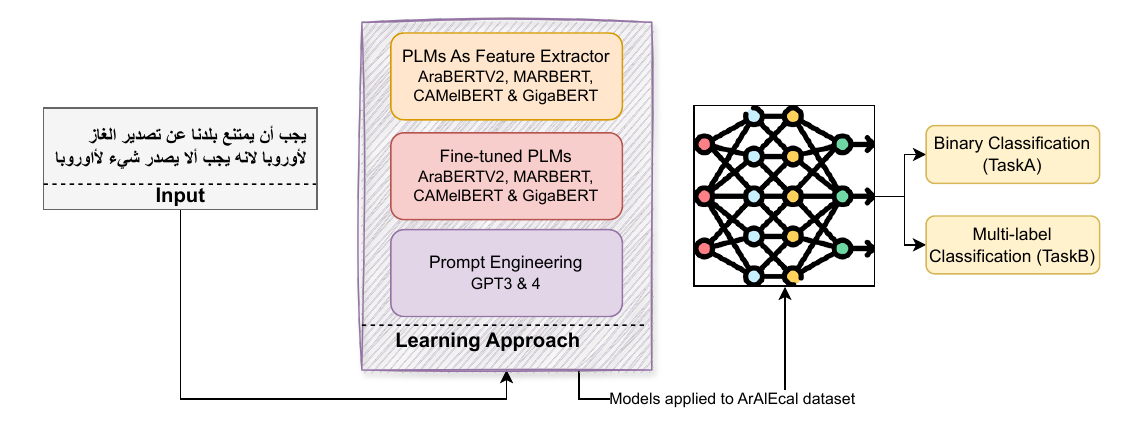}
    \caption{A visualization of our framework in detecting persuasive techniques.}
    \label{fig:framework}
\end{figure*}

Our research methodology uses the ``ArAlEval'' dataset described in greater details in Hasanain et al.~\cite{hasanain2023araieval}. The dataset consists of two collections: the firs is sourced from Arabic news sources' Twitter accounts, as documented in the previous edition of the shared task~\cite{alam2022overview}, and the second is sourced from news articles retrieved from the AraFacts dataset~\cite{ali2021arafacts}, which categorize claims verified by Arabic fact-checking websites, each claim paired with web pages either affirming or refuting it. Task 1A of our investigation entails a binary classification, aiming to identify whether a given text snippet incorporates any persuasion technique, while Task 1B is a multi-label classification, focusing on identifying the specific type of propaganda techniques employed within the text. By employing these two tasks, our study systematically delves into the persuasive tactics prevalent in Arabic social media content, fostering a deeper comprehension of influence dynamics and persuasion strategies within digital communication realms.


\begin{table}[h]
\caption{Data Statistics.}
\label{datasets_stat}
\centering
\begin{tabular}{llll}
\toprule
Binary Task (1A)      & Train       & Dev        & Test       \\ \midrule
True                  & 1918 (79\%) & 202 (78\%) & 331 (66\%) \\
False                 & 509 (21\%)  & 57 (22\%)  & 172 (34\%) \\
Total                 & 2427        & 259        & 503        \\ \midrule
Multi-label (1B)      & Train       & Dev        & Test       \\ \midrule
Persuasion Techniques & 2427        & 259        & 503       \\ \bottomrule
\end{tabular}
\end{table}

In our evaluation process, we adopt a suite of metrics tailored to the specific characteristics of our tasks. For both Task 1A and Task 1B, we use Micro-F1, weighted F1, accuracy, and the Jaccard score index. We chose weighted F1 instead of macro f1-score because it accommodates the unbalanced nature of the ArAlEval data. Accuracy is utilized for the binary classification task (i.e., Task 1A), quantifying the proportion of correctly classified instances. Meanwhile, the Jaccard score index, applied to the multi-label classification task (i.e., Task 1B), which measures the similarity between true and predicted label sets. Table~\ref{datasets_stat} presents an overview of the instance distribution across tasks. In particular, the binary classification task exhibits different proportions of True and False labels within the training, development, and test sets. Additionally, it is important to note that the multi-label classification task involves 24 labels, and the list of these labels can be found in Hasanain et al.~\cite{hasanain2023araieval}.




    

\section{Methodology}
We illustrate our framework in Figure~\ref{fig:framework}, which consists of multiple Stage:

\begin{itemize}
    \item \textbf{The input stage} involves preparing the text data that will be  analyzed for persuasive techniques. 


    \item \textbf{Learning Approach} involves adopting the types of learning that will be used for identifying different persuasive techniques in text. Our choices are based on deep learning models, which will be discussed in greater details in Section~\ref{approaches_}.

    \item \textbf{Output Stage} involves modeling the two tasks discussed in Section~\ref{Data and Task Settings}. The two tasks are binary classification of the existence or absence of persuasive techniques and multi-label classification of the specific types of persuasive techniques employed in the text. 
    
\end{itemize}


\subsection{Pre-training, fine-tuning, prompt engineering}\label{approaches_}

\textbf{Pre-training} involves training a model on a large corpus of unlabeled text data. The primary 
objective during pre-training is to allow the model to learn general language patterns, grammar, and contextual relationships within sentences.BERT (Bidirectional Encoder Representations from Transformers) is a pre-training model. BERT stands out due to its bidirectional approach, where it simultaneously considers context from both left and  right directions. During pre-training, BERT engages in self-supervised learning tasks, such as predicting  masked words in a sentence, generating its own training data from the unlabeled corpus. The model is built  upon the Transformer architecture, utilizing multiple layers of Transformer blocks to capture contextual  relationships effectively. BERT produces contextual embeddings, representing words in context and  enabling a nuanced understanding of language. Additionally, after pre-training, BERT can be fine-tuned  for specific tasks, adapting its knowledge to excel in tasks like text classification or named entity  recognition. 

 \begin{figure}[h]
    \centering
    \includegraphics[width=1\linewidth]{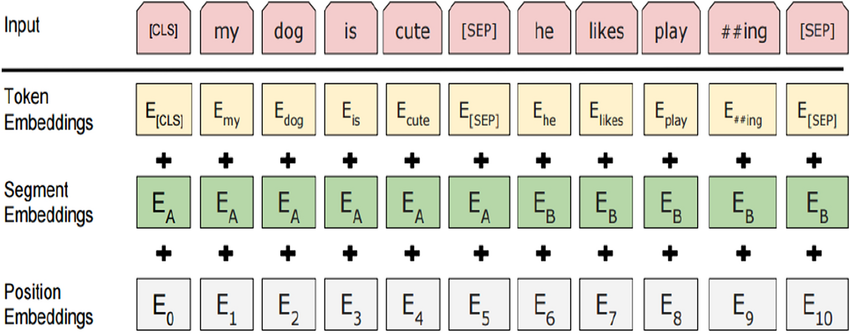}
    \caption{BERT Input representation. Taken from \cite{devlin2019bert}.}
    \label{fig:BERT}
\end{figure}

In figure \ref{fig:BERT}, the initial token in each sequence is the special classification token
[CLS]. The model utilizes the token [SEP] to distinguish between sentences. In 
instances of sentence pairs, a segment embedding is added which indicates whether the token belongs to 
sentence A or sentence B. The input representation for a particular token is formed by combining the 
associated token, segment, and position embeddings. The summation of these embeddings yields the final 
input embeddings. 
In fine-tuning, after developing the pre-trained model, which has been trained on a large dataset for 
a general task such as next token prediction and natural language understanding. Tt is now possible to begin the process of fine-tuning by updating to the model parameters. The primary goal of fine-tuning is to 
adapt the model to perform well on a particular target task or domain, which is a persuasive techniques detection in our case. During this phase, the model adjusts its parameters based on the labeled examples from the 
task-specific dataset, learning to make task-specific predictions and adapting its pre-trained knowledge to 
the intricacies of the target task. 



\begin{figure}[h]
    \centering
    \includegraphics[width=1\linewidth]{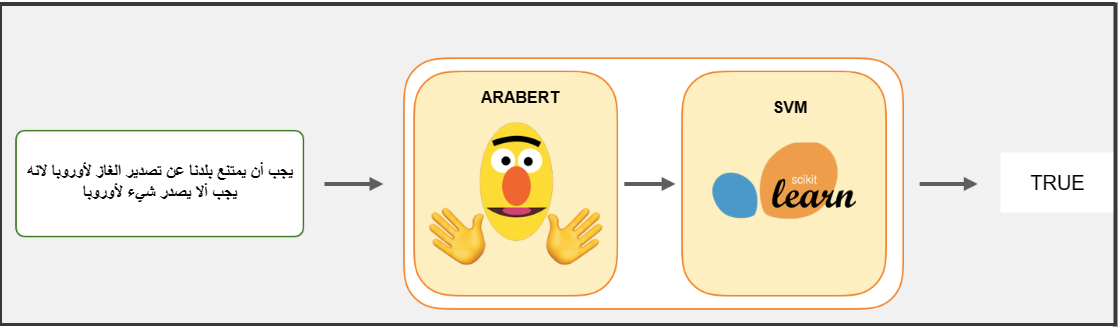}
    \caption{An illustrative example of using PLMs as feature extractor, where the extracted embeddings are passed to a Support Vector Machine (SVM).}
    \label{fig:fea-extra}
\end{figure}

\textbf{PLMs As Feature Extractors} considers the utilization of PLMs as feature extractors. In this approach, the pre-trained weights of the language models are utilized without any further fine-tuning (i.e., they are frozen). By leveraging PLMs as feature extractors, researchers can obtain contextualized embeddings for textual inputs, capturing the nuances and meaning encoded within the text. The frozen model weights ensure that the knowledge and understanding encoded in the language models are preserved, allowing researchers to focus on downstream tasks such as persuasive language detection or argument mining. This approach offers a powerful and efficient means of extracting informative features from text in case if there is a lack of GPU resources. Figure \ref{fig:fea-extra} provides an illustrative example of of this approach. The input instance is passed to a PLM, and the extracted feature vector (i.e., embedding) is used with Support Vector Machine (SVM). The benefit of this approach is that we do not have to fine-tune the model by relying on its pre-trained knowledge learned during the pre-training stage.

\textbf{Prompting engineering} plays a crucial role in guiding Artificial Intelligence (AI) models to generate outputs that align with the user's intentions. By providing specific input or instructions, the AI model can produce responses that are relevant, coherent, and in line with the desired outcomes. In this paper, we focus on leveraging the capabilities of GPT-3.5 and GPT-4 to analyze persuasive techniques employed in Arabic social media content. Let's take a closer look at the prompt we used to explore the capabilities of the GPT models. The instruction is as follows: "Classify the text according to the presence or absence of persuasive techniques. Reply with Only one word: true | false." This concise prompt was employed to guide the model in determining whether a given text exhibited persuasive techniques. The binary response of either "true" or "false" indicated the presence or absence of persuasive techniques in the text. This serves as directions for the AI models, shaping their responses and ensuring that the generated content was contextually appropriate. 



\section{Experiments}

\subsection{Implementation Details}

In our experiments examining persuaion techniques in Arabic text snippets, we utilized PyTorch \cite{NEURIPS2019_9015} for model implementation, paired with the HuggingFace platform for accessing pre-trained models such as AraBERT\cite{antoun2020arabert}, MARBERT\cite{abdul-mageed-etal-2021-arbert}, CAMelBERT\cite{inoue-etal-2021-interplay}, GigaBERT\cite{lan2020gigabert}, and AraBERTV2\cite{antoun2020arabert}. The experiments were conducted on Google Colab, providing a robust environment with GPU support. We also integrated Scikit-learn\cite{sklearn_api} for pre-processing and performance evaluations and utilized the OpenAI API to test zero-shot capabilities with GPT-3.5 and GPT-4 models\cite{openai2024gpt4, gpt-3.5/4}. It is worth mentioning that all the models utilized in this paper were exclusively tuned using the development set.

\subsection{Results}


    


    

Table~\ref{tab:results-task1a} and Table~\ref{tab:results-1b} present all scores for task1A and Task1B across all three employed metrics. A number of SOTA models are highlighted as follows: the HTE~\cite{} model leads with an F1-Micro score of 0.7634, group, MARBERT shows robust performance with scores of 0.720 F1-Micro and 0.711 F1-Weighted, illustrating moderate efficacy when large language models are employed as features without additional tuning. Significant enhancements are observed in the "Fine-tuning PLMs"group, where GigaBERT achieved the highest scores of 0.865 F1-Micro and 0.861 F1-Weighted. This suggests that fine-tuning models specifically for the task can yield superior performance. Conversely, Prompting GPT models using zero-shot learning with GPT-3.5 and GPT-4 obtained the lowest scores , indicating that while prompt engineering is a promising direction, it might still lag in effectiveness without further refinement.


\begin{table}[h]
\caption{The results of Task1A (Binary Classification) on the test set.}
\label{tab:results-task1a}
\centering
\begin{tabular}{lccc}
\toprule
Model/Metric   & F1-Micro & F1-Weighted & Accuracy  \\ \midrule
\multicolumn{4}{c}{Previous Works} \\ \midrule
HTE     & 0.763 &  ---  & ---  \\
KnowTellConvince & 0.757 & --- & --- \\
 rematchka & 0.755 & --- & --- \\
 \midrule
\multicolumn{4}{c}{PLMs As Feature Extractors} \\ \midrule
AraBERT     & 0.662 & 0.656   & 0.503  \\
MARBERT       & 0.720 & 0.711   & 0.564  \\
CAMelBERT  & 0.684 & 0.663   & 0.517  \\
GigaBERT  & 0.710 & 0.706    & 0.555 \\ \midrule
\multicolumn{4}{c}{Fine-tuning PLMs}  \\ \midrule
AraBERTV2   & 0.838 & 0.814    & 0.716 \\
MARBERT   & 0.780 & 0.683    & 0.609 \\
CAMelBERT   & 0.819 & 0.798    & 0.691 \\
GigaBERT  & 0.865 & 0.861   & 0.768\\ \midrule
\multicolumn{4}{c}{Prompt Engineering PLMs}   \\ \midrule
GPT-3.5 (ZShot)  & 0.525    & 0.537       & 0.525    \\
GPT4 (ZShot)   & 0.531    & 0.543       & 0.531    \\
\bottomrule 
\end{tabular}
\end{table}

Table~\ref{tab:results-1b} showcases the performance of various models on Task1B, a multi-label classification task. In the segment for "Previous Works," UL \& UM6P lead with an F1-Micro score of 0.566, suggesting a strong baseline. Using PLMs as feature extractors, CAMelBERT shows a relative advantage with an F1-Micro score of 0.442 and a Jaccard index of 0.30, indicating its adeptness at handling the multi-label complexity better than its counterparts. The "Fine-tuning PLMs" group revealed further enhancements, with GigaBERT achieving the top F1-Micro and Jaccard scores of 0.532 and 0.464, respectively. This illustrates the benefits of model-specific tuning for this task. However, the "Prompt Engineering PLMs" approach, utilizing zero-shot techniques with GPT-3.5 and GPT-4, performs significantly lower, with GPT-4 only reaching an F1-Micro of 0.258 and a Jaccard of 0.191, highlighting the challenges and limitations of applying generalist models to specialized tasks without extensive customization.

Tables~\ref{tab:results-task1a} and~\ref{tab:results-1b} provide a comprehensive overview of model performances for Task 1A (binary classification) and Task 1B (multi-label classification) respectively. In both tables, a clear trend emerges where fine-tuning PLMs yields superior results compared to using PLMs solely as feature extractors or employing prompt engineering techniques. For Task 1A, GigaBERT stands out as the top-performing model, achieving high scores across all metrics. Conversely, in Task 1B, GigaBERT also leads after fine-tuning, demonstrating its versatility across different classification tasks. However, while prompt engineering with zero-shot learning techniques show potential, they generally lag behind in performance compared to fine-tuned PLMs. These observations demonstrates the importance of model optimization and task-specific adaptation in achieving optimal performance in natural language processing tasks.
Prompt engineering in Task 1A (binary classification) showed decent results for both GPT-3.5 and GPT-4 models. the effectiveness of prompt engineering depends on the specific task. While promising for binary classification, it might not translate directly to tasks requiring more complex classification as we will see for task1B. In the next section, we performed few-shot learning to evaluate effectiveness of prompt engineering, showing potential directions for further research. 



\begin{table}[h]
\caption{The results of Task1B (Multi-label Classification) on the test set.}
\label{tab:results-1b}
\centering
\begin{tabular}{lccc}
\toprule
Model/Rep                 & F1-Micro & F1-Weighted & Jaccard \\ \midrule
\multicolumn{4}{c}{Previous Works} \\ \midrule
 UL \& UM6P & 0.566 &  ---  & ---  \\
 rematchka & 0.565 &  ---  & ---  \\
 AAST-NLP & 0.552 &  ---  & ---  \\
 \midrule
\multicolumn{4}{c}{ PLMs As Feature Extractors}                \\ \midrule
AraBERT                   & 0.413    & 0.415       & 0.282   \\
MARBERT                   & 0.393    & 0.389       & 0.259   \\
CAMelBERT                 & 0.442    & 0.432       & 0.300   \\
GigaBERT                  & 0.416    & 0.409       & 0.276   \\ \midrule
\multicolumn{4}{c}{Fine-tuning  PLMs}                       \\ \midrule
AraBERTV2                 & 0.493    & 0.4267      & 0.422   \\
MARBERT                   & 0.459    & 0.380       & 0.373   \\
CAMelBERT                 & 0.502    & 0.417       & 0.421   \\
GigaBERT                  & 0.532    & 0.445       & 0.464   \\ \midrule
\multicolumn{4}{c}{Prompt Engineering  PLMs}                       \\ \midrule
GPT-3.5 (ZShot) & 0.275    & 0.299       & 0.222   \\
GPT4 (ZShot)          & 0.258    & 0.294       & 0.191  \\ \bottomrule
\end{tabular}
\end{table}

Finally, previously highlighted models, that examined the same dataset as the one used in this paper, achieved higher performance compared to our approaches for task 1B. We observed that this difference in performance can be attributed to the utilization of specific loss functions that address the imbalanced nature of the data labels and account for multi-label classification scenarios. This observation aligns with the findings of Alhuzali and Ananiadou~\cite{alhuzali2021spanemo}, who discovered that employing a loss function tailored to multi-label cases significantly improved performance. However, it is important to note that our current paper did not prioritize achieving state-of-the-art (SOTA) results. Instead, our focus was on exploring a range of NLP methods and utilizing a suite of powerful PLMS techniques. While our performance may not have surpassed previous works, our objective was to gain insights and investigate various approaches within the field.

\section{Analysis}

\subsection{Evaluation of the results of Task1A and Task1B}


\begin{figure*}[h]
  \centering
  \begin{minipage}[b]{0.45\linewidth}
    \centering
    \includegraphics[width=\linewidth]{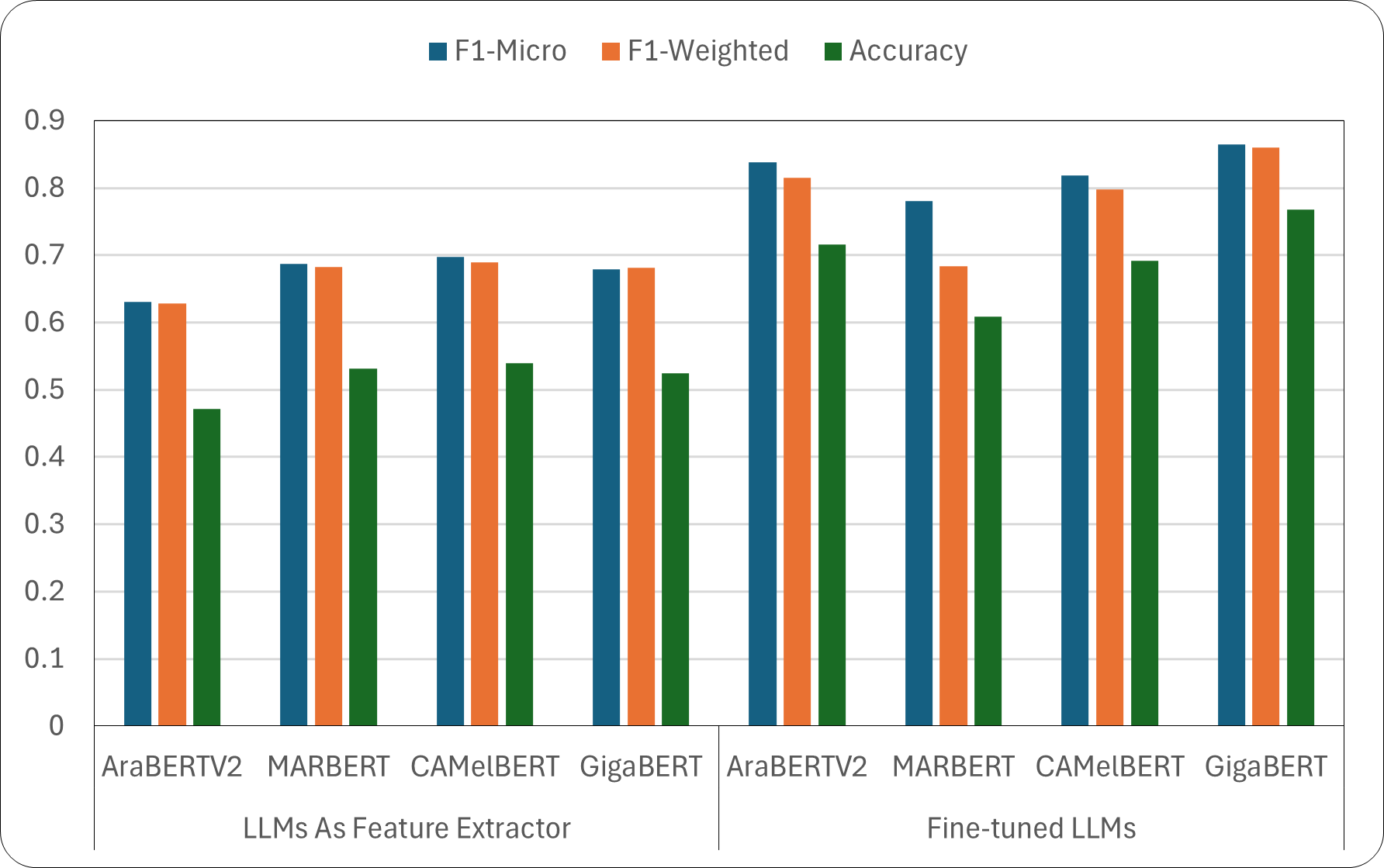}
    \captionsetup{justification=centering}
    \caption*{\textbf{(a)} Results of Task 1A}
    \label{fig:task1a}
  \end{minipage}
  \hfill
  \begin{minipage}[b]{0.5\linewidth}
    \centering
    \includegraphics[width=\linewidth]{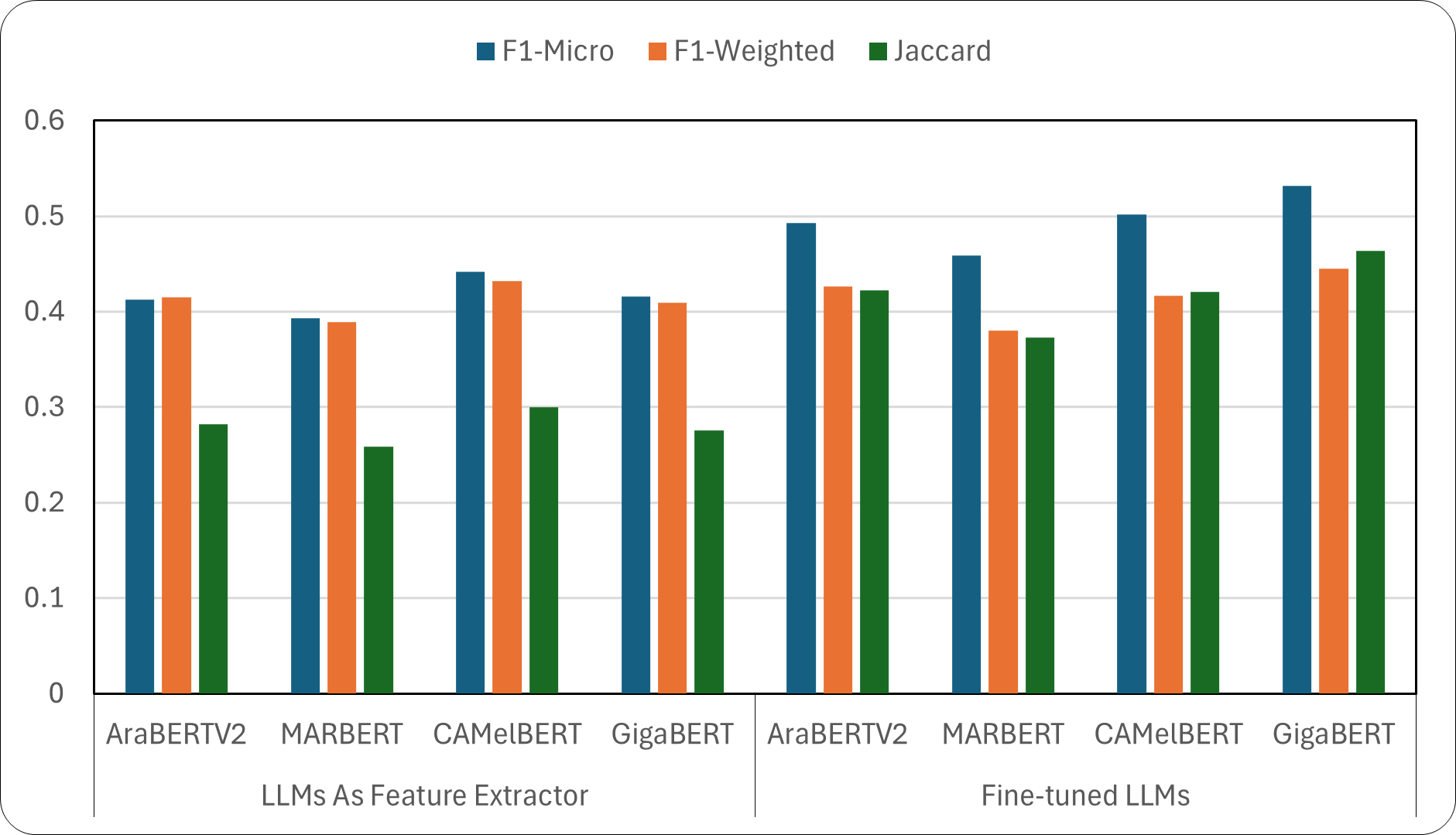}
    \captionsetup{justification=centering}
    \caption*{\textbf{(b)} Results of Task 1B}
    \label{fig:task1b}
  \end{minipage}
  \caption{The results of two approaches, i.e., PLMs as Feature extraction and Fine-Tuned PLMs.}
  \label{fig:task_comparison}
\end{figure*}

Figure~\ref{fig:task_comparison} presents a visual of the results for both Task 1A and Task 1B, comparing the performance of two approaches. The first approach, referred to as "PLMs as feature extractors," includes AraBERTV2, MARBERT, CAMelBERT, and GigaBERT. Among these models, CAMelBERT demonstrates slightly superior results compared to MARBERT and GigaBERT. In contrast, the second approach, termed "fine-tuned PLMs," evaluated the same set of models that have been fine-tuned on the target task. Notably, the fine-tuned models in the second approach outperformed the PLMs as feature extractors in the first approach. GigaBERT, in particular, achieved the highest performance among these fine-tuned models. This suggests that fine-tuning the models on the specific task yields better results, demonstrating the potential of adapting these models to the task and domain of interest. By fine-tuning the models, they become more specialized and capable of achieving higher performance on the targeted task.


The findings depicted in Figure \ref{fig:task_comparison} shed light on the results of Task 1B. Interestingly, this figure reveals similar patterns to those observed in the one observed for Task 1A, albeit with slightly lower performance. This discrepancy can be attributed to the inherent complexity of Task 1B, which involved multi-label classification. The consistent observation across both figures demonstrated the importance of customizing models for the specific task at hand. By tailoring the models to the task of interest, they gain the ability to acquire in-depth knowledge about the task and adjust their weights accordingly. 



\subsection{Effect of Few-shot Learning}
\begin{figure}[h!]
\centering
\includegraphics[width=0.9\linewidth]{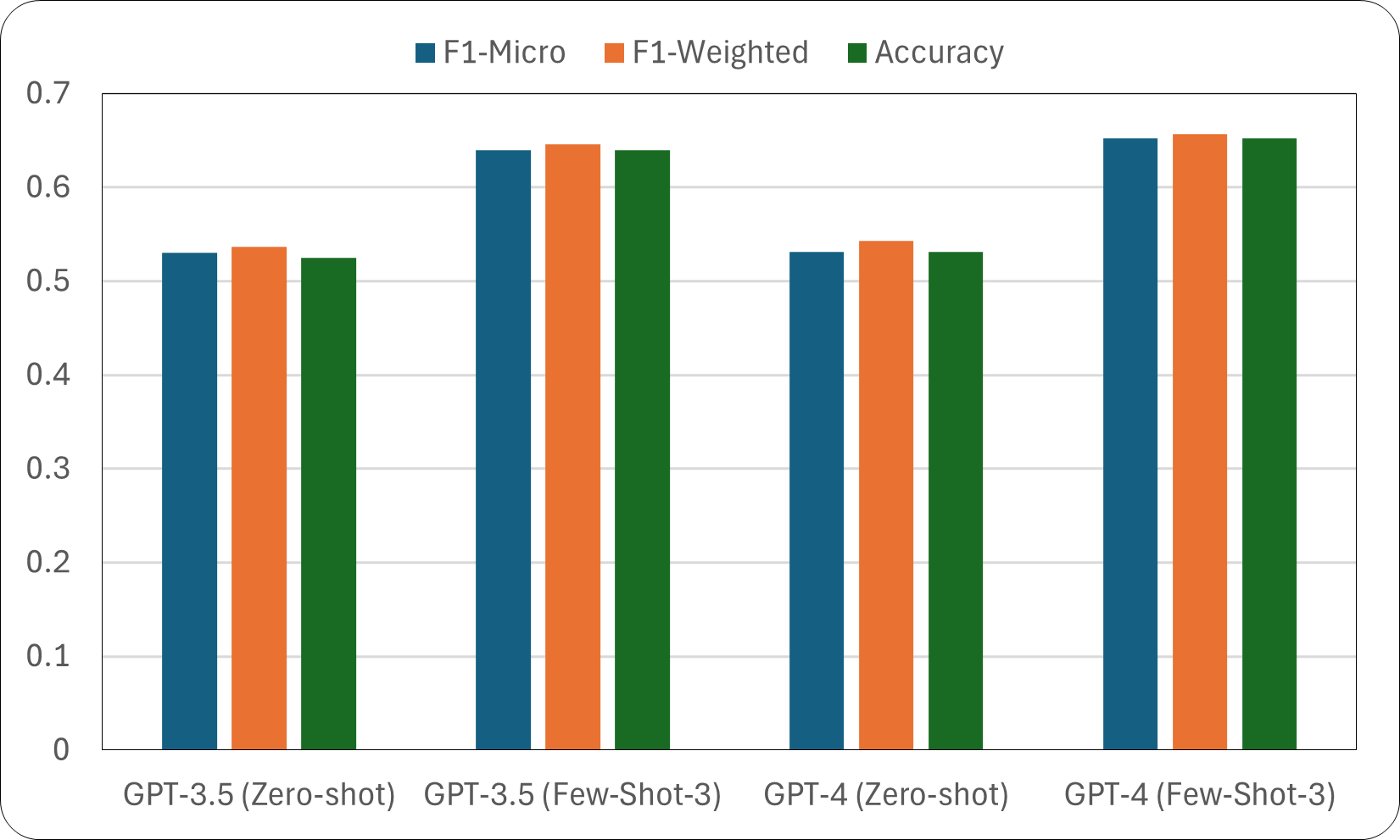}
\caption{Evaluation of Zero-shot vs Few-shot capabilities of GPT models.} 
\label{fig:gpt_ana}
\end{figure}

In this analysis, we examined the effectiveness of GPT models in few-shot settings as shown in Figure~\ref{fig:gpt_ana}, where only a limited number of instances were available. Specifically, we evaluated the performance of GPT-3.5 and GPT-4 on Task1A, which involves binary classification. Initially, we assessed the models in a zero-shot setting, where they were not fine-tuned or trained on any specific data for the task. Surprisingly, the results indicate that both GPT-3.5 and GPT-4 achieve very low scores compared to approaches involving fine-tuning or feature extraction. This suggests that without any prior task-specific training, the models struggle to perform well. 

However, when we transition to the few-shot setting, an interesting shift occurs. Remarkably, the GPT models exhibit improved performance compared to the zero-shot scenario, even with the utilization of only three labeled instances. Thus, the evaluation of GPT-3.5 and GPT-4 in the few-shot setting sheds light on their adaptability and learning capabilities, further emphasizing their potential to excel with limited data.









\section{Conclusion}
This study aimed to address the pervasive issue of identifying and counteracting persuasive techniques disseminated across social media. Drawing upon the ArAIEval persuasion technique dataset, which encompasses both binary and multi-label classifications, our investigation is focused on three distinct learning methodologies. Our experiments demonstrated the critical role of fine-tuning Pre-trained Language Models (PLMs), with GigaBERT emerging as an exemplar performer across all evaluation metrics. This highlighted the indispensable nature of tailoring models to specific tasks, as fine-tuning substantially boosted performance for persuasive technique detection. 

Looking forward, our study highlighted promising avenues for future inquiry. Particularly compelling is the exploration of prompt engineering within few-shot settings, where preliminary findings showcased notable performance improvements. By refining and expanding upon the capabilities of few-shot learning techniques, we stand to unlock even greater performance gains, especially in scenarios characterized by limited labeled data. Furthermore, investigating the efficacy of specialized loss functions tailored to multi-label classification scenarios holds potential for mitigating challenges associated with imbalanced data labels, thereby augmenting classification accuracy and resilience. In short, our research lays a solid foundation for understanding persuasive technique detection on social media, setting the stage for advancements in combating the proliferation of persuasive misinformation in digital spaces.

\section*{CRediT authorship contribution statement}
\textbf{Abdurahman Alzahrani}: Implementation of Prompt engineering  and Writing - Introduction \& Related work. \textbf{Eyad Babkier}: Implementation of Prompt engineering  and Writing - Data and task settings, Implementation details \& Results. \textbf{Firas Yanbaawi}: Implementation of fine-tuning, and Writing - Abstract, Introduction, Methodology.
\textbf{Faisal Yanbaawi}: Implementation of using PLMs as feature extraction, Writing - Introduction, Related Works, and analysis. \textbf{Hassan Alhuzali}: Supervision, Writing – review \& editing.

\section*{Acknowledgment}
We extend our appreciation to Google Colab for enabling us to conduct the experiments presented in this paper. 

\textbf{Declaration of generative AI and AI-assisted technologies in the writing process:} During the preparation of this work, we used ChatGPT, an AI chatbot developed by OpenAI, in order to improve our written work. After using this tool, we reviewed and edited the content as needed and take full responsibility for the content of the publication.

\bibliographystyle{IEEEtran}
\bibliography{main}

\begin{IEEEbiography}[{\includegraphics[width=1in,height=1.25in,clip,keepaspectratio]{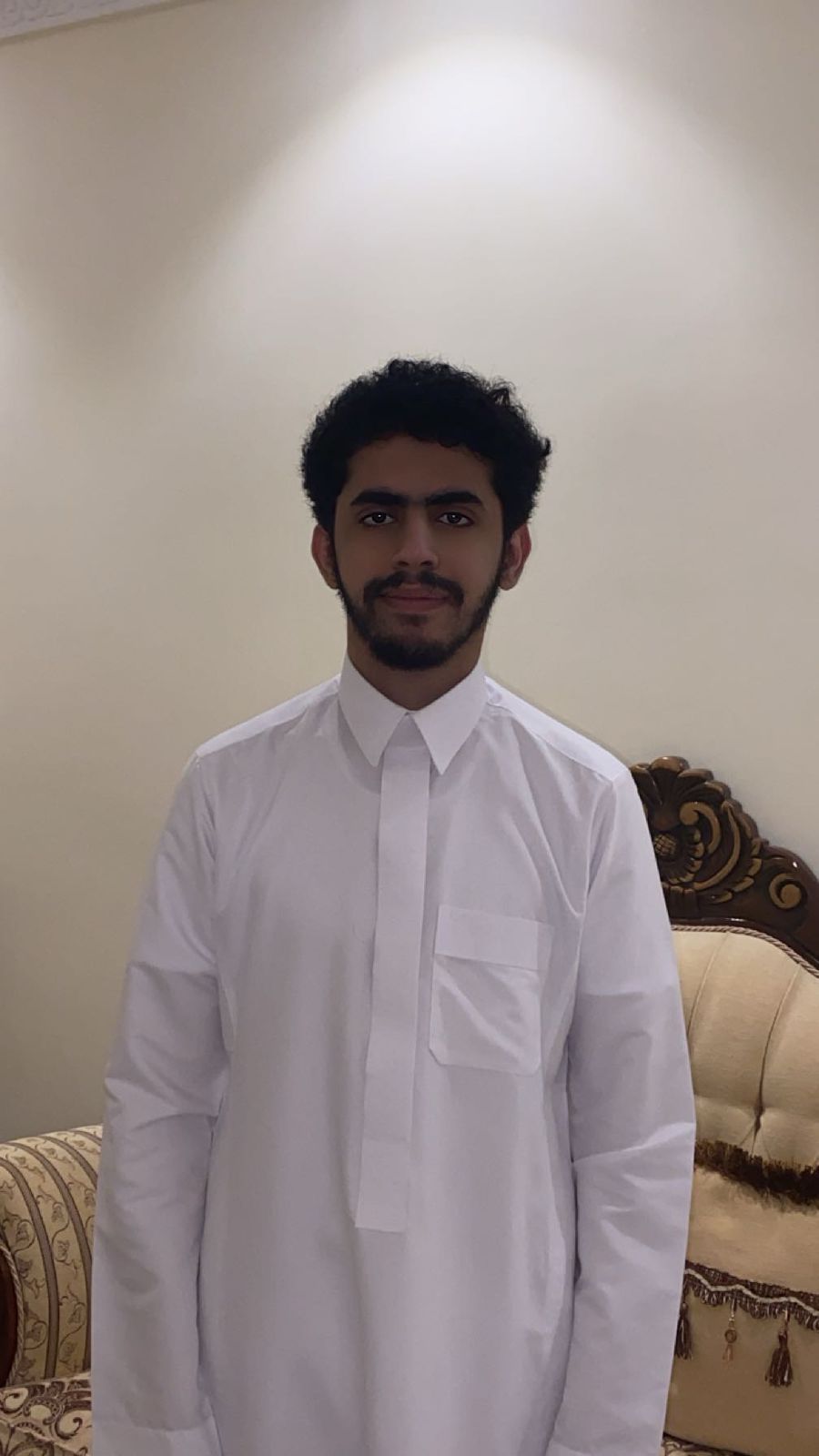}}]{AbdulRahman Alzahrani} is in his fourth year at Umm Al-Qura University, studying Computer Science and Artificial Intelligence, set to graduate next semester. Alongside his degree, he's taken specialized courses like AI chatbots and smart learning. His understanding of NLP and persuasion techniques grew through his graduation project under the supervision of Dr. Hassan Alhuzali, enhancing both his scholarly skills and his grasp of NLP's practical applications and persuasion's societal impact.
\end{IEEEbiography}

\begin{IEEEbiography}[{\includegraphics[width=1in,height=1.25in,clip,keepaspectratio]{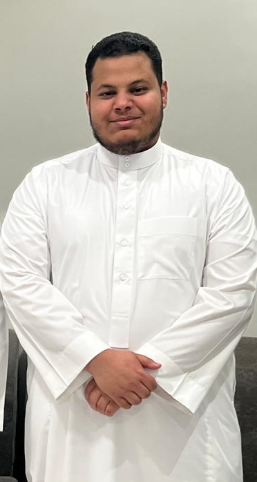}}]{Eyad Babkier} is currently in his last semester at Umm Al-Qura University, majoring in Computer Science and Artificial Intelligence. Throughout his degree, he has learned skills in Front-End Development, Data Analysis, and Machine Learning. His graduation project, supervised by Dr. Hassan Alhuzali, focused on Natural Language Processing (NLP) and persuasive techniques. This project significantly enhanced his technical abilities and provided deep insights into the practical applications of NLP and the impact of persuasive techniques on individuals and society.

\end{IEEEbiography}

\begin{IEEEbiography}[{\includegraphics[width=1.5in,height=1.5in,clip,keepaspectratio]{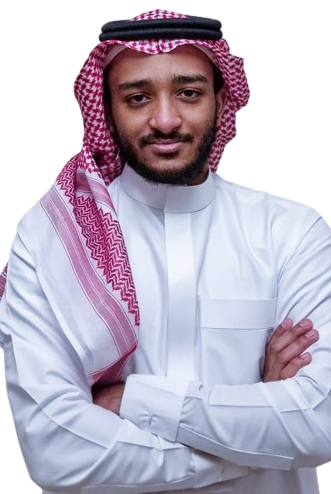}}]{Faisal Yanbaawi} is in his fourth year at Umm Al-Qura University, studying Computer Science and Artificial Intelligence. Alongside his degree, he's taken specialized courses like web development, data analysis and machine learning. His understanding of NLP and persuasion techniques grew through his graduation project under the supervision of Dr. Hassan Alhuzali, enhancing both his scholarly skills and his grasp of NLP's practical applications and persuasion's societal impact.

\end{IEEEbiography}

\begin{IEEEbiography}[{\includegraphics[width=1in,height=1.25in,clip,keepaspectratio]{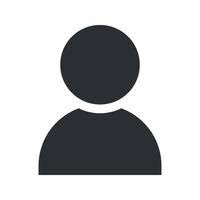}}]{Firas Yanbaawi} is currently in his final semester pursuing a Bachelor's degree in Computer Science and Artificial Intelligence at Umm Al-Qura University. Alongside his degree, he's taken specialized courses like web development, data analysis and machine learning. His understanding of NLP and persuasion techniques grew through his graduation project under the supervision of Dr. Hassan Alhuzali, enhancing both his scholarly skills and his grasp of NLP's practical applications and persuasion's societal impact.
\end{IEEEbiography}

\begin{IEEEbiography}[{\includegraphics[width=1in,height=1.25in,clip,keepaspectratio]{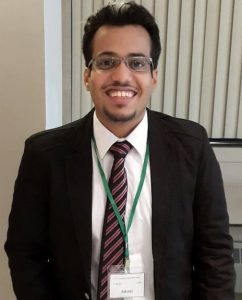}}]{HASSAN ALHUZALI} obtained his Ph.D. degree in Computer Science from the University of Manchester, UK in 2022. Prior to that, he served as an Associate Researcher at the University of Manchester. In 2016, he earned an M.S. degree in Information Science from Indiana University-Bloomington, USA. Following the completion of his master’s degree, he embarked on a period as a visiting student at the Positive Psychology Center at UPENN, USA, as well as UBC, CA. Currently, he serves as an Assistant Professor at Umm Al-Qura University in SA. His ongoing research focuses on natural language processing, affective computing, and mental health.
\end{IEEEbiography}

\EOD

\end{document}